\documentclass[10pt,twocolumn,letterpaper]{article}
\pdfoutput=1
\usepackage{cvpr}
\usepackage{times}
\usepackage{epsfig}
\usepackage{graphicx}
\usepackage{amsmath}
\usepackage{amssymb}
\usepackage{subfigure}
\usepackage{hyperref}
\hypersetup{breaklinks=true,letterpaper=true,colorlinks}

\cvprfinalcopy 

\def\cvprPaperID{****} 

\ifcvprfinal\fi
\begin{document}

\title{Attention-based Joint Detection of Object and Semantic Part}

\author{
Keval Morabia\\
{\tt\small morabia2@illinois.edu}
\and
Jatin Arora\\
{\tt\small jatin2@illinois.edu}\\
University of Illinois at Urbana-Champaign\\
\and
Tara Vijaykumar\\
{\tt\small tgv2@illinois.edu}
}

\maketitle

\begin{abstract}
    In this paper, we address the problem of joint detection of objects like dog and its semantic parts like face, leg, etc. Our model is created on top of two Faster-RCNN models that share their features to perform a novel Attention-based feature fusion of related Object and Part features to get enhanced representations of both. These representations are used for final classification and bounding box regression separately for both models. Our experiments on the PASCAL-Part 2010 dataset show that joint detection can simultaneously improve both object detection and part detection in terms of mean Average Precision (mAP) at IoU=0.5.
\end{abstract}

\section{Introduction}
With the advances in camera and imaging technology, the amount of available online content with images and videos is ever-increasing. This makes it impossible for any human to manually parse and make sense out of images and motivates the development of automated techniques to parse images. One such popular and immensely important task is automated object detection in images. Recently, there have been major developments in this domain (\cite{yolov3}, \cite{ren2015faster}, \cite{he2017mask}) but the diversity of this problem poses new challenges motivating active research in this direction.

Humans identify objects not only by looking at the object as a whole, but looking at various parts of the object to gain confidence. Taking inspiration from this idea, there have been some recent researches (\cite{felzenszwalb2009object}, \cite{yao2018exploiting}). A concrete example is, identifying a human in an image by identifying parts like hands, faces etc. to add confidence to classification. Another example is identifying an aeroplane by using information of its wings. \cite{yao2018exploiting} propose a LSTM-based architecture for joint modelling of semantic-part information with full object detection. Recently, self-attention based transformer architectures \cite{vaswani2017attention} have gained popularity which eliminated the use of Recurrent Model like RNN for modeling contextual information. We aim to borrow ideas of self-attention in our setting where semantic parts associated with an object will act as different contexts which will be weighted based on the attention score.

In this project, we propose a novel model for joint detection of an object as a whole and its semantic parts using an attention-based feature fusion of object and its related parts. We use the PASCAL-Part 2010 Dataset (\cite{chen2014detect}) that contains segmentation masks for object and its various semantic parts. From our limited experiments for Animal objects (bird, cat, cow, dog, horse, sheep) and part (face, leg, neck, tail, torso, wings) detection, we find that our joint detection model can give slight improvement for object detection and a reasonable improvement for part detection in terms of mean Average Precision (mAP) at Intersection over Union (IoU) threshold of 0.5. We believe a thorough hyperparameter tuning can lead to even better results.

\section{Model Architecture}
As a starting point, we use pyTorch's Faster-RCNN implementation \footnote{https://github.com/pytorch/vision/blob/master/torchvision/models/detection/} (with Resnet50 backbone and Feature Pyramid Network) for object detection. Then we will use another Faster-RCNN model for semantic part detection which will be trained only to identify parts and not objects.

\begin{figure*}[t]
\includegraphics[width=\textwidth]{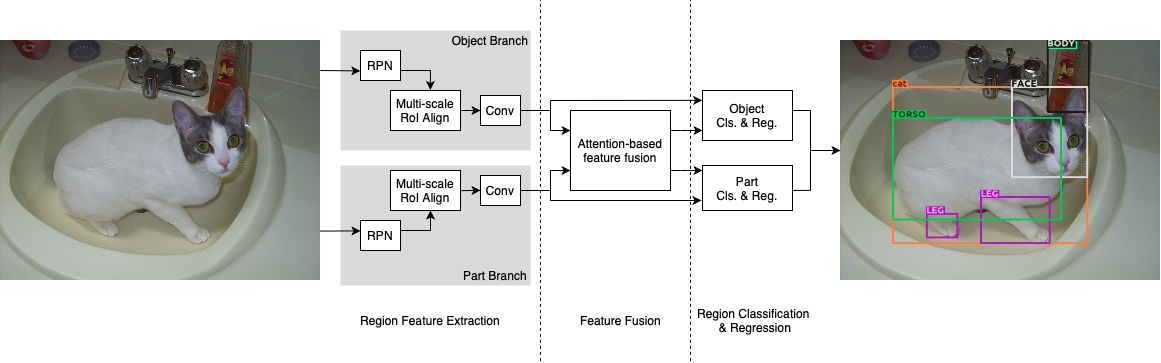}
  \caption{Architecture for Joint Detection of Object and Semantic Part using Attention-based feature fusion}
  \label{fig:architecture}
\end{figure*}

We believe that both these object detection and part detection model performances can be improved by using information of the other in a joint training architecture. Our model architecture is highly motivated from \cite{chen2014detect} in that we replace the Relationship modeling and LSTM-based feature fusion with an Attention-based feature fusion architecture. We divide our model architecture in 3 parts: (i) Region Feature Extraction, (ii) Attention-based Feature Fusion, and (iii) Region Classification \& Regression as shown in Fig:\ref{fig:architecture}.

\begin{figure*}[t]
    \centering
    \subfigure[]{\includegraphics[width=6cm]{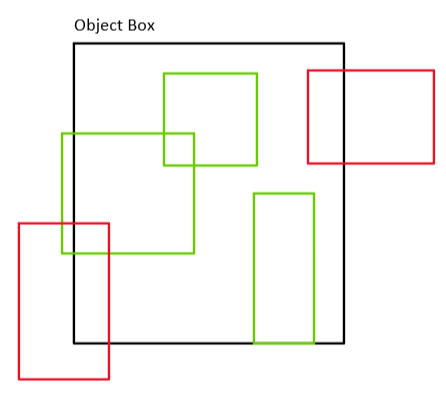}}
    \hspace{30pt}
    \subfigure[]{\includegraphics[width=6cm]{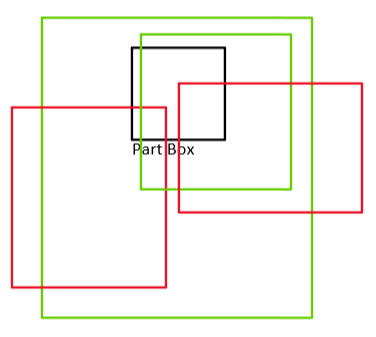}}
    \caption{(a) Example of related parts for an object. (b) Example of related objects for a part. Here green boxes are selected for fusion (thresh=0.9) where as red boxes are not selected since the intersection area of object and part is less than fusion\_thresh * part\_area for red boxes}
    \label{fig:fusion_eg}
\end{figure*}

\subsection{Region Feature Extraction}
In this part, we take the image as input and pass it to 2 different Region Proposal Networks (RPN) \cite{ren2015faster}, one for object and another for parts, and produce object and part proposals respectively. From the feature map produced by the RPNs, we apply Multi-Scale RoI Align with a Feature Pyramid Network to get a fixed sized feature for each object/part proposal.

\subsection{Attention-based Feature Fusion}

Once we have all the object and part proposals from the previous step, we get an enhanced representation of object and part proposals that can be used in next step for final object and part detection. To get this new features for object and part boxes, we perform an attention-based feature fusion for object (or parts) and its related parts (or objects). We define a hyperparameter called $fusion\_thresh (f)$ that decides which object and part proposals boxes are related to each other and should undergo fusion. We consider those object and part boxes where: \\
$intersection\_area(object, part) \geq f \times area(part) $ \\
Fig:\ref{fig:fusion_eg} shows an example of related objects (or parts) that will be used for fusion to get fused part representation of related parts (or related objects).

Once we have related parts (or objects) for an object (or part), we learn an attention layer that gives a score for each object-part pair. This score will be used to get weighted average of features of all related parts (or objects) for the object (or part). This weighted average is the enhanced surrounding part (or object) representation for an object (or part). Note that the reason why we are weighing each related parts (or objects) for an object (or part) is because not all related object part pairs are relevant, for example a person can be standing next to a car as well, and not all nearby objects (or parts) are equally important.  We expect this to give better performance than naive average of related parts (or objects).

\subsection{Region Classification \& Regression}
In this step, we use the object features concatenated with fused related part features for final classification and bounding box regression for objects. Similarly, we use the part features concatenated with fused related object features for final classification and bounding box regression for parts.



\begin{figure*}[t]
\centering
\includegraphics[width=10cm]{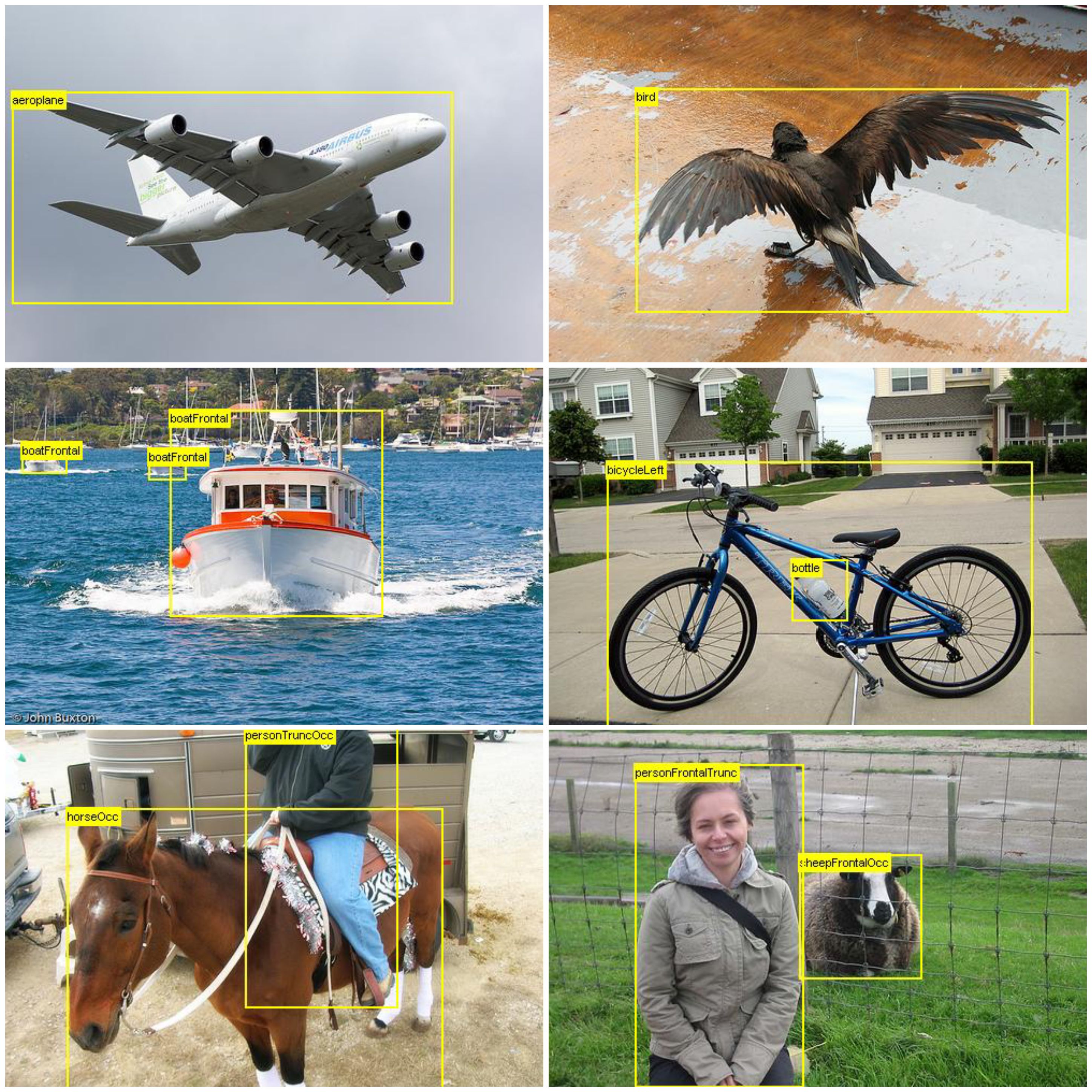}
  \caption{Example images and ground truth object bounding boxes from PASCAL VOC 2010 dataset}
  \label{fig:example_imgs}
\end{figure*} 

\begin{figure*}
\includegraphics[width=\textwidth]{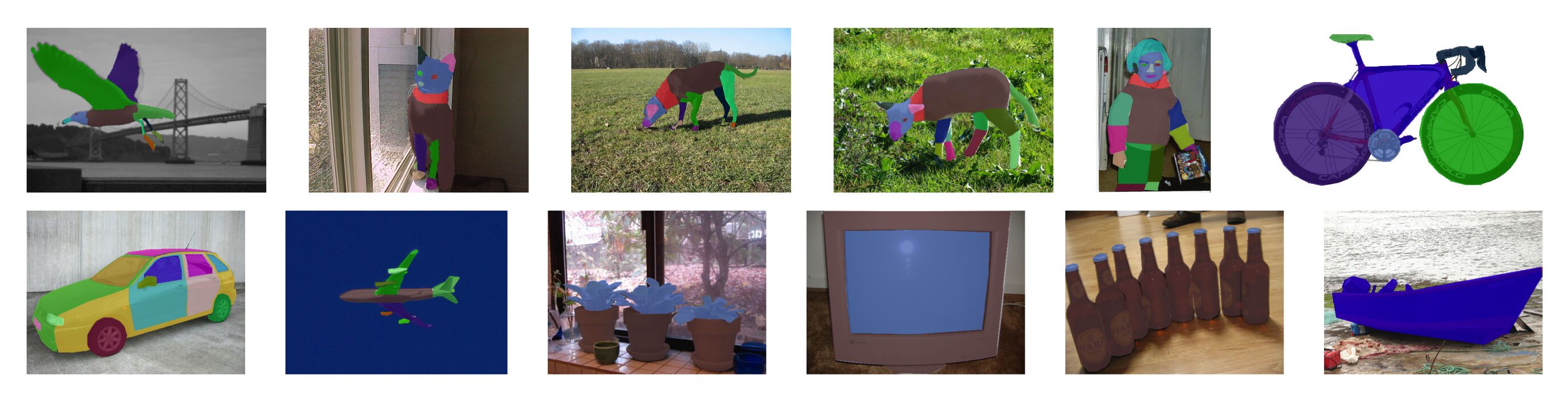}
  \caption{Example segmentation annotations from the PASCAL-Part Dataset}
  \label{fig: part_masks}
\end{figure*}

\section{PASCAL-Part Dataset 2010}

\begin{figure*}[t]
\centering
\includegraphics[width=15cm]{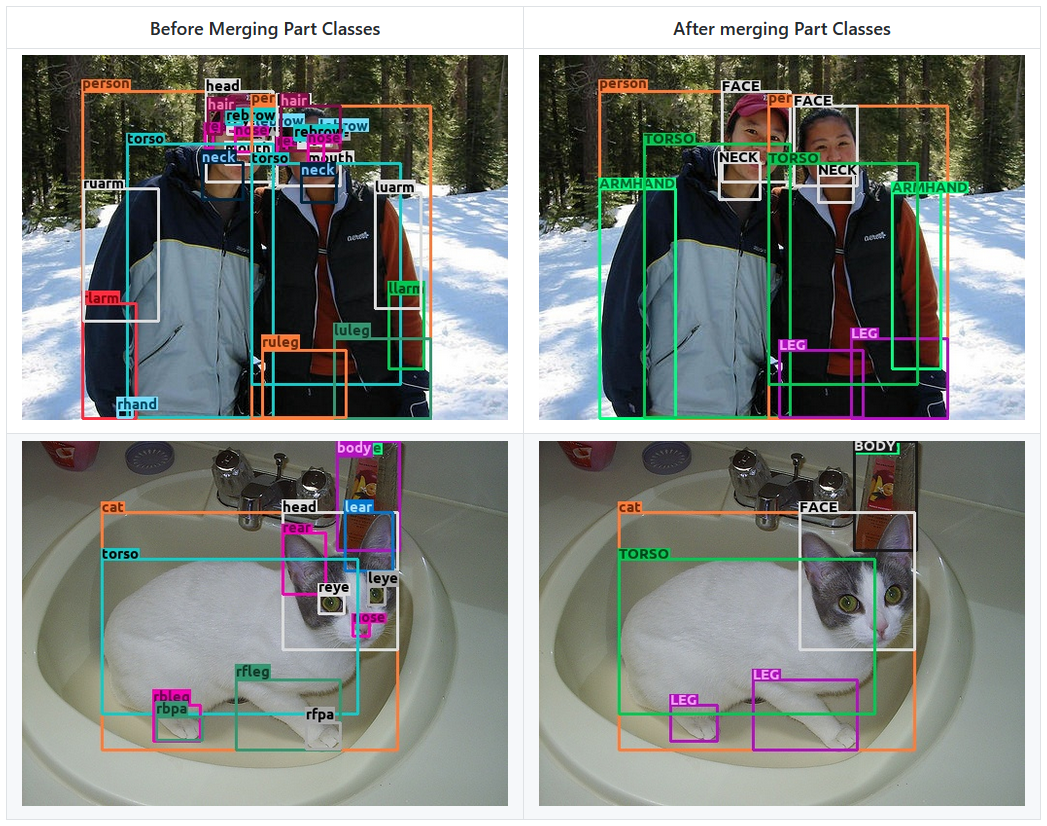}
  \caption{Example of merged smaller part classes into a coarse-grained part class for PASCAL-Part Dataset}
  \label{fig: merged_parts}
\end{figure*}

PASCAL VOC 2010 dataset \footnote{http://host.robots.ox.ac.uk/pascal/VOC/voc2010/} is a popular dataset used to benchmark Object Detection models in which each image has an annotation file containing bounding box coordinates and class labels for each object as shown in Fig:\ref{fig:example_imgs}. There are 20 classes present in the dataset which can be categorized into 4 super categories namely Person, Animal, Vehicle, Indoor.  Training and validation contain 10,103 images while testing contains 9,637 images. The twenty object classes in Pascal VOC 2010 dataset are:
\begin{enumerate}
    \item Person: person
    \item Animal: bird, cat, cow, dog, horse, sheep
    \item Vehicle: aeroplane, bicycle, boat, bus, car, motorbike, train
    \item Indoor: bottle, chair, dining table, potted plant, sofa, tvmonitor
\end{enumerate}

\begin{table}
\centering
	\begin{tabular}{|c|c|c|}
	\hline
	\textbf{Category} & \textbf{\# Train Samples} & \textbf{\# Validation Samples}\\
	\hline
	\texttt{person} & {1714} & {1829}\\
    \hline
    \texttt{animals} & {1872} & {1886}\\
    \hline
    \texttt{vehicles} & {1493} & {1483}\\
    \hline
    \texttt{indoor} & {692} & {697}\\
    \hline
	\end{tabular}
\caption{Overall category-wise distribution in PASCAL-Part Dataset}
\label{Table:category_wise_distribution}
\end{table}

The PASCAL-Part Dataset (\cite{chen2014detect}) is a set of additional annotations for PASCAL VOC 2010 as shown in Fig:\ref{fig: part_masks}. It provides segmentation masks for each body part of the object and silhoutte annotation for categories that do not have a consistent set of parts (eg: boat). The part annotations were originally given in .mat format which we preprocessed using the scipy module to convert it into .json format. The annotations only contain segmentation masks, so we use the rectangular box containing this segmentation as the bounding box coordinates that are used for Object Detection. This dataset can also be used for animal part detection and general part detection. Note that these part annotations are only given for train and val images.

    
    

\subsection{Dataset Statistics}

\begin{table}[t]
\centering
	\begin{tabular}{|c|c|}
	\hline
	\textbf{Class} & \textbf{\# Samples}\\
	\hline
	\texttt{person.head} & {3960}\\
    \hline
    \texttt{person.lear} & {937}\\
    \hline
    \texttt{horse.lear} & {213}\\
    \hline
    \texttt{cow.tail} & {91}\\
    \hline
    \texttt{car.door\_3} & {1}\\
    \hline
	\end{tabular}
\caption{Uneven part distribution in raw training data}
\label{Table:training_part_distribution}
\end{table}

\begin{table}[t]
\centering
    \begin{tabular}{|c|c|c|}
    \hline
    \textbf{Category} & \textbf{\# Train Samples} & \textbf{\# Validation Samples}\\
    \hline
    \texttt{horse} & {208} & {316}\\
    \hline
    \texttt{dog} & {706} & {708}\\
    \hline
    \texttt{cat} & {563} & {568}\\
    \hline
    \texttt{bird} & {484} & {484}\\
    \hline
    \texttt{sheep} & {350} & {358}\\
    \hline
    \texttt{cow} & {233} & {239}\\
    \hline
    \end{tabular}
\caption{Class-wise object distribution for animals in dataset}
\label{Table:obj_wise_animal_distribution}
\end{table}

\begin{table*}[t]
\centering
	\begin{tabular}{|c|c|c|}
	\hline
	\textbf{Model} & \textbf{Animal Object Detection} & \textbf{Animal Part Detection}\\
	\hline
	\texttt{Object Detection} & {87.2} & {--}\\
    \hline
    \texttt{Part Detection} & {--} & {51.3}\\
    \hline
    \texttt{Joint Object and Part Detection} & {\textbf{87.5}} & {\textbf{52.0}}\\
    \hline
	\end{tabular}
\caption{mean Average Precision at Intersection over Union threshold 0.5 for Animal object and part detection on PASCAL-Parts validation dataset}
\label{Table:results}
\end{table*}

\begin{table}[t]
\centering
  \begin{tabular}{|c|c|c|}
  \hline
  \textbf{Category} & \textbf{\# Train Samples} & \textbf{\# Validation Samples}\\
  \hline
  \texttt{FACE} & {2933} & {2909}\\
  \hline
  \texttt{LEG} & {5217} & {5324}\\
  \hline
  \texttt{NECK} & {1628} & {1651}\\
  \hline
  \texttt{TAIL} & {1148} & {1163}\\
  \hline
  \texttt{TORSO} & {2978} & {2974}\\
  \hline
  \texttt{WINGS} & {189} & {175}\\
  \hline
  \end{tabular}
  \caption{Class-wise part distribution for animals in dataset}
\label{Table:part_wise_animal_distribution}
\end{table}

For our study, we use the original train-validation splits provided with the dataset. The training data has \textbf{4,998} images and validation set has \textbf{5,105} images. Table:\ref{Table:category_wise_distribution} presents the category-wise distribution of object classes in the training set.

For part detection, the training dataset has very fine-grained labels spanning across 166 different classes, with many classes not having enough training samples. Some stats of the distribution of classes in the training data are shown in Table:\ref{Table:training_part_distribution}. To address the uneven distribution of part labels and for faster experimentation, we work on only objects in the animals category, and collate fine-grained object-part classes into coarser ones, for example, \texttt{FACE} includes \texttt{beak, hair, head, nose, lear, lebrow, leye, mouth, rear, rebrow, reye}. Similarly, \texttt{WINGS} includes \texttt{lwing, rwing}.

Since, out overall task is to jointly model object and part detection, we remove those samples from consideration where for an object, the part information is not available. In Table: \ref{Table:part_wise_animal_distribution}, we present the distribution of part samples in training and validation sets for the animals category which we are working with.

\section{Evaluation Metric}
Object Detection models are evaluated using the metric called mean Average Precision (mAP). There are many components involved in computing mAP, which are Average Precision, and Intersection over Union (IoU). IoU measures the overlap between ground truth boundaries and predicted boundaries for an object. A prediction is considered correct if IoU $>$ threshold. Average Precision (AP) measures the average of precision values where recall values ranges from 0 to 1. mAP is the mean of AP values for all different classes in the dataset. mAP values are generally reported at IoU threshold of 0.5, but it can also be reported at IoU thresholds [0.5, 0.55, 0.6, ..., 0.95] and taken average of all these mAP values.

\section{Experiments}

\begin{table}[t]
\centering
	\begin{tabular}{|c|c|c|}
	\hline
	\textbf{Category} & \textbf{Object Detection} & \textbf{Joint Detection}\\
	\hline
	\texttt{horse} & {88.6} & {88.6}\\
    \hline
    \texttt{dog} & {82.0} & {88.8}\\
    \hline
    \texttt{cat} & {85.7} & {92.6}\\
    \hline
    \texttt{bird} & {84.9} & {86.7}\\
    \hline
    \texttt{sheep} & {87.6} & {84.8}\\
    \hline
    \texttt{cow} & {92.4} & {83.0}\\
    \hline
	\end{tabular}
\caption{Class-wise object detection AP at IOU=0.5 for animals in dataset}
\label{Table:object_ap}
\end{table}

\begin{table}[t]
\centering
	\begin{tabular}{|c|c|c|}
	\hline
	\textbf{Category} & \textbf{Part Detection} & \textbf{Joint Detection}\\
	\hline
	\texttt{FACE} & {25.9} & {83.3}\\
    \hline
    \texttt{LEG} & {83.7} & {53.8}\\
    \hline
    \texttt{NECK} & {53.6} & {33.0}\\
    \hline
    \texttt{TAIL} & {33.0} & {32.1}\\
    \hline
    \texttt{TORSO} & {30.6} & {80.3}\\
    \hline
    \texttt{WINGS} & {81.1} & {29.3}\\
    \hline
	\end{tabular}
\caption{Class-wise part detection AP at IOU=0.5 for animals in dataset}
\label{Table:part_ap}
\end{table}

For training all models, we use the initial model parameters that are pretrained on MS COCO dataset \footnote{http://cocodataset.org/}, which is much larger than PASCAL VOC 2010 dataset and contains about 4 times the number of classes, and replace the last classification and regression layers to make predictions for objects and parts. This transfer learning helps model converge training very quickly in less number of training epochs and achieve better performance. While training, we randomly flip images horizontally with probability 0.5 as a data augmentation technique to make the model more robust.

\begin{figure*}[t]
    \centering
    \subfigure[]{\includegraphics[width=8.5cm]{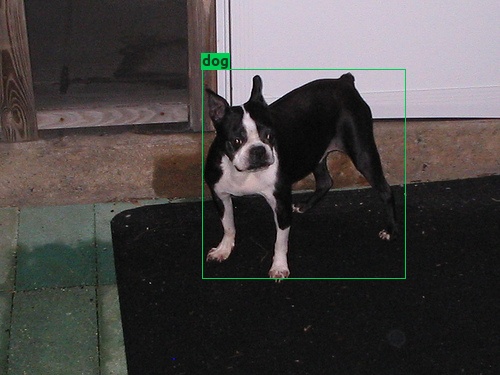}}
    \hspace{5pt}
    \subfigure[]{\includegraphics[width=8.5cm]{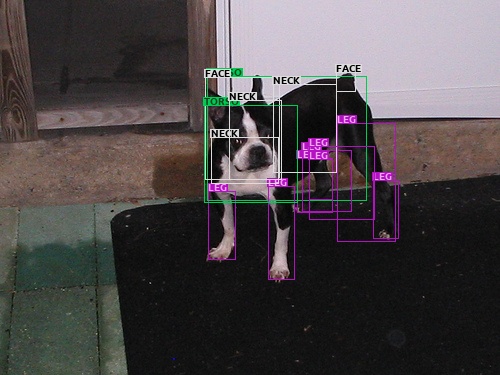}}
    
    \subfigure[]{\includegraphics[width=8.5cm]{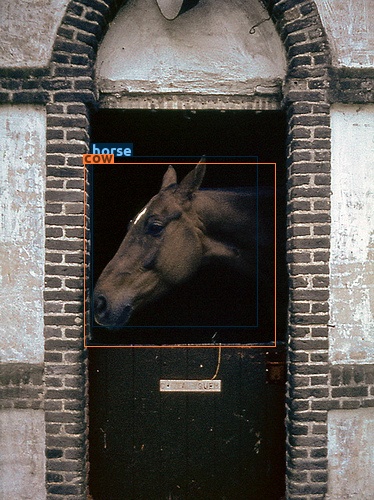}}
    \hspace{5pt}
    \subfigure[]{\includegraphics[width=8.5cm]{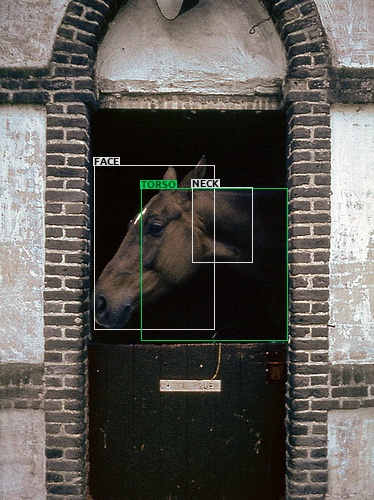}}
    \caption{Sample outputs from our joint trained model. For cleanliness, we have labelled objects and parts separately.  }
    \label{fig:qualitative_results}
\end{figure*}

\begin{figure*}[t]
    \centering
    \subfigure[]{\includegraphics[width=8.5cm]{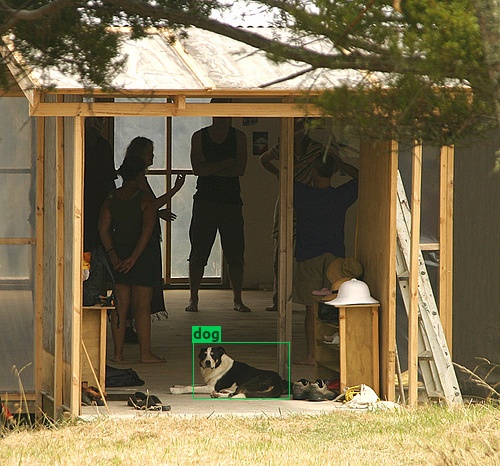}}
    \hspace{5pt}
    \subfigure[]{\includegraphics[width=8.5cm]{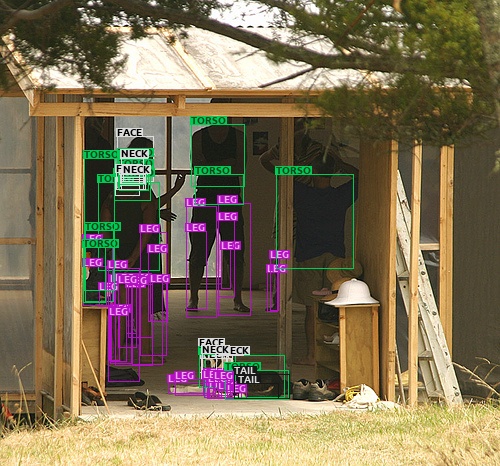}}
    \caption{Example output where person parts are also detected when training and evaluating only for animal classes}
    \label{fig:part_issue}
\end{figure*}

The joint detection model has about 82m parameters and training takes 12 minutes per epoch. We train the model for 15 epochs using SGD optimizer with initial learning rate 1e-3 (decayed by 0.1 after every 5 epochs), momentum 0.9, weight decay 1e-6, batch size of 1 image, and a fusion threshold of 0.9. Note that for VOC2010, no annotated test data is available so all our results are evaluated on PASCAL VOC 2010 val dataset. For faster experiments, we train and report results only on Animal object (bird, cat, cow, dog, horse, sheep) and part (face, leg, neck, tail, torso, wings) classes. As shown in Table:\ref{Table:results}, the joint detection model gives slight improvement in terms of mAP@IoU=0.5 for animal object detection, and reasonable improvement for animal part detection. Table:\ref{Table:object_ap} and \ref{Table:part_ap} provide a detailed analysis of Average Precision for each of the object and part classes. We believe a thorough hyperparameter tuning can lead to even better results. In Figure \ref{fig:qualitative_results}, we present some sample outputs from our jointly trained model. In (b) we can observe that although the parts of the dog are detected correctly, there are multiple overlapping detections. We present more details on the part detection performance and our experiments with non-max suppression to handle this, in the next subsection.

Note: We also experimented with a naive average of related parts (or objects) instead of weighted average using attention scores for joint detection, but that did not lead to any improvement in performance over 2 separate object and part detection models.

\subsection{Discussion on Part Detection performance}
As it can be seen in Table:\ref{Table:results}, part detection performance for animal part classes is very less as compared to object detection performance for single part detection model as well as Joint detection model. The reason for this is because in animals sub-dataset that we used for all experiments, there are many persons as well. Just looking at a leg, the model might get confused between person leg and animal leg and hence would detect both legs. Hence, there will be lots of false part detections for animals class. This can be observed below in Fig:? where person leg and torso are also detected even when the annotations used for training are only for animals objects. We expect this issue to be absent when we train the model for detection of all object and part classes.

\textbf{Varying Part detection NMS inference threshold}:
We observed that there is a large number of overlapping part detections and there is high variance in the areas of their detected bounding boxes. This may be a cause of reduced mAP scores. Hence, we tried varying the NMS threshold for part evaluation of our Joint Detection model and present our finding in Table:\ref{Table:part_nms} showing no improvement in performance with varying the threshold.
\begin{table}
\centering
	\begin{tabular}{|c|c|c|}
	\hline
	\textbf{NMS Threshold} & \textbf{mAP @I0U=0.5}\\
	\hline
	\texttt{0.1} & {49.1}\\
    \hline
    \texttt{0.3} & {51.0}\\
    \hline
    \texttt{0.4} & {51.9}\\
    \hline
    \texttt{0.45} & {52.1}\\
    \hline
    \texttt{0.5} & {52.0}\\
    \hline
    \texttt{0.55} & {51.6}\\
    \hline
	\end{tabular}
\caption{NMS Threshold Variation for Part evaluation of Joint Detection Model}
\label{Table:part_nms}
\end{table}

\section{Future Work}
We expected more improvement from the Joint detection model for object and part detection. Hence, we plan to investigate the feature fusion part deeply. We also plan to create attention score visualizations to see if the scores are coming out to be as expected or not. For the above results, we performed all experiments on animal object and part classes, but we also want to see results on all object and part classes. Finally, we will also tune hyperparameters thoroughly for better performance.

\section{Links}
\begin{enumerate}
    \item \textbf{Code}: \href{http://github.com/kevalmorabia97/Object-and-Semantic-Part-Detection-pyTorch}{github.com/kevalmorabia97/Object-and-Semantic-Part-Detection-pyTorch}
    \item \textbf{Video Presentation}: \href{https://tinyurl.com/joint-detection}{tinyurl.com/joint-detection}
\end{enumerate}

{\small
\bibliographystyle{ieee_fullname}
\bibliography{joint-detection}
}

\end{document}